\documentclass[letterpaper, 10 pt, conference]{ieeeconf}  

\IEEEoverridecommandlockouts                              

\usepackage{graphicx} 
\usepackage{amsmath} 
\usepackage{amssymb}  
\usepackage{array}
\usepackage{soul}
\usepackage{hyperref}

\usepackage{siunitx}
\usepackage{color}
\newcommand{\tgli}[1]{{ \color{black}#1}}

\newcommand{\kefi}[1]{{ \color{black}#1}}

\title{\LARGE \bf
VLM-TDP: VLM-guided Trajectory-conditioned Diffusion Policy for Robust Long-Horizon Manipulation
}

\author{Kefeng Huang$^{1*}$, Tingguang Li$^{1*}$, Yuzhen Liu$^{1 \dag}$, Zhe Zhang$^{2}$, Jiankun Wang$^{2}$, Lei Han$^{1\dag}$
\thanks{* Equal contribution.}
\thanks{$\dag$ {Corresponding author.}}
\thanks{$^{1}$ Kefeng Huang, Tingguang Li, Yuzhen Liu, Lei Han are affiliated with Tencent Robotics X, China, (email: {\tt\footnotesize{\{kefhuang, teaganli, rickyyzliu, lxhan\}}@tencent.com})}%
\thanks{$^{2}$ Zhe Zhang, Jiankun Wang are affiliated with Southern University of Science and Technology in Shenzhen, China, email: {\tt\footnotesize{wangjk}@sustech.edu.cn})
        }%
}

\begin{document}

\maketitle
\thispagestyle{empty}
\pagestyle{empty}


\begin{figure*}[thpb]
    \centering
    \includegraphics[width=\textwidth]{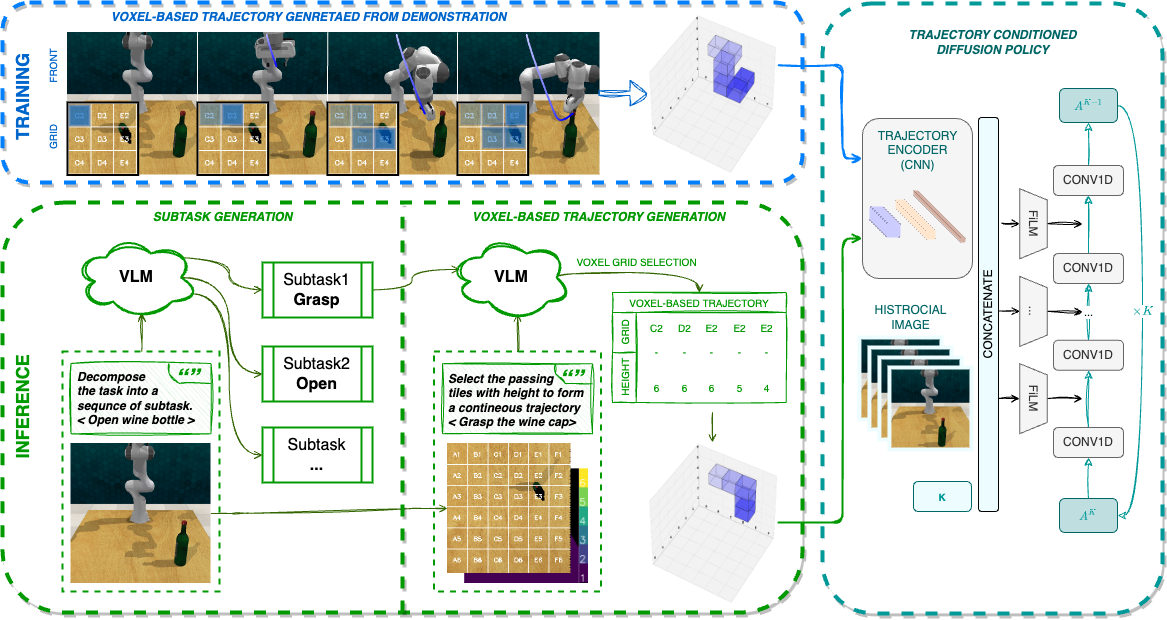}
    \caption{The overview of VLM-TDP. We utilize VLM to decompose a task into a series of sub-tasks and then generate trajectories for each sub-task (bottom left). The generated trajectories are then fed into a trajectory-conditioned diffusion policy which achieves robust performance(right). During training, the trajectory was extracted from the demonstration (top left).}
    \label{fig:tdp}
    \vspace{-0.5cm}
\end{figure*}


\begin{abstract}
Diffusion policy has demonstrated promising performance in the field of robotic manipulation. However, its effectiveness has been primarily limited in short-horizon tasks, and its performance significantly degrades in the presence of image noise.  
To address these limitations, we propose a VLM-guided trajectory-conditioned diffusion policy (VLM-TDP) for robust and long-horizon manipulation.
Specifically, the proposed method leverages state-of-the-art vision-language models (VLMs) to decompose long-horizon tasks into concise, manageable sub-tasks, while also innovatively generating voxel-based trajectories for each sub-task.
The generated trajectories serve as a crucial conditioning factor, effectively steering the diffusion policy and substantially enhancing its performance.
The proposed Trajectory-conditioned Diffusion Policy (TDP) is trained on trajectories derived from demonstration data and validated using the trajectories generated by the VLM.
\kefi{Simulation} experimental results indicate that our method significantly outperforms classical diffusion policies, achieving an average 44\% increase in success rate, over 100\% improvement in long-horizon tasks, and a 20\% reduction in performance degradation in challenging conditions, such as noisy images or altered environments. \kefi{These findings are further reinforced by our real-world experiments, where the performance gap becomes even more pronounced in long-horizon tasks. Videos are available on \url{https://youtu.be/g0T6h32OSC8}}
\end{abstract}

\section{Introduction}
In recent years, robot manipulation has emerged as a popular research area, driven by the growing demand for autonomous systems capable of performing diverse tasks in dynamic and unstructured environments. 
With advancements in machine learning, researchers are increasingly turning to imitation learning for robot manipulation tasks.
This approach allows robots to learn directly from human demonstrations, bypassing the need for manual control algorithms \cite{mandlekar2021matters, cui2022play, florence2019self}.

Among various imitation learning methods,
diffusion policies \cite{chi2023diffusion, ze20243d} have shown significant promise by approximating action distributions using denoising diffusion probabilistic models (DDPM) \cite{ho2020denoising}, especially in multimodal and high-dimensional action spaces. Recognizing the potential of this approach, several efforts have been made to enhance the classical diffusion policy.
For instance, \cite{ze20243d} modifies the input observations to point clouds to improve performance, while \cite{ke20243d} introduces 3D relative transformers and language conditioning in a multi-task setup. \cite{li2024language} also incorporates language and object conditioning to enhance capabilities. Despite these advancements, the majority of existing works are limited to short-horizon tasks, and their strong dependence on single-modal visual inputs  makes them vulnerable to noise and input variability, particularly in complex and dynamic tasks.



In parallel, Large Language Models (LLMs) and Vision-Language Models (VLMs) trained on Internet-scale data, have gained attention in the field of robotics \cite{achiam2023gpt, radford2021learning,  chowdhery2023palm}. 
Prior works have shown the ability of large models to break down complex instructions provided in natural language into actionable task plans \cite{ahn2022can, chen2023open, huang2022language} and perform visual reasoning \cite{chen2024spatialvlm, yang2023set}. 
Specifically, \cite{ahn2022can} demonstrated the ability of VLMs to select pre-trained skills, while \cite{liang2023code} \cite{huang2023voxposer} showed the effectiveness in generating robot policy code. Additionally, recent research by \cite{liu2024moka} revealed that VLMs can generate affordances and produce 2D trajectories through a grid-tile selection process, indicating their potential to create trajectories based on multi-choice prompts.


Drawing inspiration from these advancements, we propose a VLM-guided trajectory-conditioned diffusion policy for robust and long-horizon manipulation. 
We begin our approach by employing VLM to decompose long-term tasks into manageable short-term sub-tasks. 
Compared to complex, long-horizon tasks, these sub-tasks are more suitable for execution via diffusion policies, which substantially improve the overall system performance. 
In addition, for each sub-task, we utilize VLM to effectively generate 3D voxel-based spatial trajectories. These trajectories are subsequently used as extra conditioning modalities to control the diffusion network, enabling it to generate robot actions that accomplish the manipulation tasks. 

\kefi{
To validate the effectiveness of our proposed voxel-based trajectory representation, we evaluated our method across multiple simulation benchmarks and real-world tasks. The simulation experiments were conducted using the RLBench \cite{james2020rlbench} and the Colosseum\cite{pumacay2024colosseum}, while real-world evaluations were performed on a Franka Panda robot. Our method demonstrates significant advantages in both effectiveness and robustness compared to language-conditioned and conventional diffusion policies. Through extensive experiments, we observe that VLM-TDP achieves a 44\% improvement in average task success and a 100\% improvement in long-horizon tasks, with consistent results across simulation and real-world evaluations. The robustness of our approach is evidenced by a 20\% smaller performance degradation under varying or noisy environmental conditions, which we attribute to the incorporation of spatial trajectory modality. 
}

Overall, the contributions of this paper can be summarized as follows:
\begin{itemize}
    \item We propose an innovative approach that leverages VLMs to decompose complex, long-horizon tasks into multiple concise, manageable sub-tasks.  These sub-tasks are individually addressed and then combined, resulting in a significantly enhanced performance when tackling long-horizon tasks.
    \item We propose a novel voxel-based spatial trajectory generation method utilizing VLMs. These generated trajectories are seamlessly incorporated as an additional crucial conditioning modality for diffusion policies, significantly enhancing the system robustness and success rate.
\end{itemize}

\section{Related Works}

\subsection{Diffusion Models in Robotics}
Diffusion models~\cite{sohl2015deep, ho2020denoising} are generative models that gradually transform random noise into samples from a target distribution using probabilistic frameworks. Owing to their notable success in image generation~\cite{ho2020denoising,song2020denoising,song2020score,rombach2022high}, diffusion models have been extensively used for data augmentation~\cite{yu2023scaling,mandi2022cacti,chen2023genaug}, expanding training datasets and enhance diversity. Additionally, they have been applied to modify the policy observation~\cite{chen2024dreamarrangement}, where a diffusion model generates a goal image conditioned on the task description, which is then provided to the planner policy.

Another application of diffusion models is in direct robot control. One of the popular approaches is diffusion policy~\cite{chi2023diffusion}, where the diffusion model predicts an action sequence conditioned on observations. This approach has demonstrated advantages in handling multimodal action distributions and high-dimensional action spaces. A series of works~\cite{bartsch2024sculptdiff, ze20243d, li2024language} has further explored this direction.
Another prominent variation involves the use of diffusion models to denoise the end-effector trajectory, followed by a low-level planner to execute the planned poses~\cite{ma2024hierarchical,ke20243d}.

\subsection{Policy Conditioning Representation}
Policy conditioning is a powerful tool when adapting a well-performing model to a specific goal, with multiple approaches available.
A straightforward method is goal conditioning, commonly achieved through the use of a goal image~\cite{chen2024dreamarrangement,bousmalis2023robocat,lynch2020learning}. For tasks like sculpting, the goal state can be represented as a point cloud of the target shape~\cite{bartsch2024sculptdiff}. However, goal conditioning often includes irrelevant information, such as background elements or extraneous objects, and may be limited to certain task types.

Another prevalent approach is language conditioning~\cite{ke20243d,jang2022bc,brohan2023rt,nair2022learning,ahn2022can} or object conditioning~\cite{stone2023open,zhu2023learning}. \cite{li2024language} combines both approaches by using language to select the target object and then conditioning the policy on the object's point cloud.
RT-Trajectory~\cite{gu2023rt} offers an innovative approach by conditioning the policy with a trajectory. This method modifies the RT-1~\cite{brohan2022rt} framework, training the policy based on a trajectory, and demonstrates improved performance over language- or goal-conditioned policies. In contrast to our voxel-based trajectory, which captures height information, their trajectory is a continuous 2D or 2.5D line (using color to represent height), which is more difficult to generate.

\subsection{Vision-Language Models in Planning}
Large Language Models and Vision-Language Models have demonstrated impressive language comprehension, common knowledge, and reasoning abilities \cite{achiam2023gpt, radford2021learning,  chowdhery2023palm,chen2024spatialvlm, yang2023set}. This makes them highly effective in interpreting human instructions and generating structured plans for robotic tasks \cite{ahn2022can, chen2023open, huang2022language}.
For instance, \cite{ingelhag2024robotic} presented an approach that leverages these foundational models to select specific visuomotor diffusion policy skills for different tasks. However, for unfamiliar tasks, it remains necessary to collect new demonstrations and train additional policies.
Previous studies~\cite{liu2024moka} have highlighted the capabilities of VLMs, such as GPT-4V(ision)~\cite{achiam2023gpt}, in solving multiple-choice problems, often outperforming methods that directly generate continuous planned values. By utilizing annotated images with grids and textual notations, these VLMs can generate trajectories made up of waypoints on an image. 

In a concurrent study, \cite{lee2024affordance} proposed a similar method of enhancing VLM capabilities for trajectory selection. However, their approach combines trajectory selection with the identification of affordance points on objects to shape the reward function in reinforcement learning. In contrast, our method relies on the VLM solely solely for trajectory selection, while the affordance point is determined by the policy itself, guiding the diffusion policy in task execution.
Although we share a similar approach in enabling VLMs to plan in 3D space, our objectives differ. Their focus is on improving sample efficiency in reinforcement learning, whereas our goal is to enhance robust, long-horizon manipulation through diffusion policy. Combining both approaches reveals broader possibilities for voxel-based trajectory planning.

\section{Methodology}

\subsection{Framework Overview}
In this work, we address the challenges encountered by diffusion policies in accomplishing long-horizon tasks and their sensitivity to noise and input variations. Consequently, we propose employing a vision-language model to decompose complex long-horizon tasks into a series of more manageable and concise sub-tasks. For each individual sub-task, the vision-language model generates an end-effector trajectory, effectively guiding the diffusion policy towards task completion. We hypothesize that a diffusion policy, when conditioned on both RGB and trajectory information, exhibits a higher level of resilience to variations, as compared to policies purely conditioned on RGB information.
An overview of our framework is shown in Fig.~\ref{fig:tdp}.
In the following sections, we first introduce the task decomposition and trajectory generation module in Section~\ref{sec:vlm_plan}, and the trajectory conditioned diffusion policy in Section~\ref{sec:diffusion}.

\subsection{Task Decomposition and Trajectory Generation via VLM} \label{sec:vlm_plan}

We employ the VLM in two distinct stages: task decomposition and end-effector trajectory generation. The task decomposition stage breaks down a complex task into a series of sub-tasks, while the trajectory generation stage generates detailed trajectories for each individual sub-task.

\begin{figure}
    \centering
    \includegraphics[width=\linewidth]{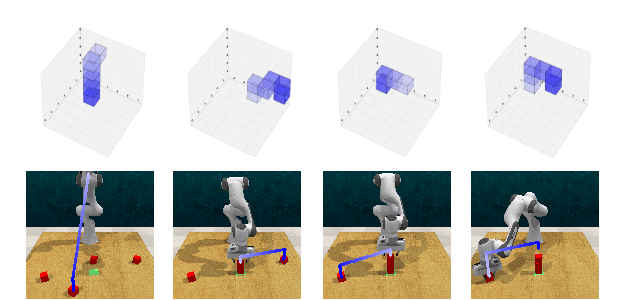}
    \caption{Examples of voxel-based spatial trajectory and their corresponding end-effector movement (blue lines). }
    \label{fig:subtask}
    \vspace{-0.5cm}
\end{figure}

\textbf{Task Decomposition.}
The objective of this stage is to decompose the complex task into several shorter, manageable sub-tasks. We define a sub-task as a discrete phase of the manipulation task that begins with the opening or closing the gripper and ends with closing or opening the gripper, typically indicating the robot's complete interaction with an object. 
This division of tasks eliminates the occurrence of repeated or reversed trajectories within a sub-task, thereby minimizing potential ambiguity in the trajectory generation process later.
Given a high-level task description $\mathcal{I}$ and an initial visual observation $O_0$, we query a vision-language model $\mathcal{M}$ to decompose the complete task into a series of $n$ sub-tasks $\mathcal{M}(O_0, \mathcal{I}) = \{\mathcal{S}_i\}_{i=1,...n}$, where $\mathcal{S}_i$ represents a language description for each individual sub-task, which outlines the specific object and action. 
For example, in the task of \emph{Put Item in Drawer}, VLM generates 4 sub-tasks including "grasp the bottom drawer handle", "pull the drawer out", "grasp the item", and "put the item into the bottom drawer". 
These intermediate sub-task descriptions are then passed to the trajectory generation module. 

\textbf{Trajectory Generation via Mask-based Visual Prompting.}
In this module, we need VLM to generate a series of 3D coordinates given 2D images. 
The 2D images from the front camera are first projected to a top-down view. To maintain visual consistency, we fill in the shadows caused by obstructions with the table texture. Additionally, a height map is created to represent the elevation of each pixel.
Inspired by~\cite{liu2024moka}, we use a set of markers as visual prompts to enable VLM to apply its spatial reasoning capabilities to predict trajectories in 3D space. We divide the observed RGB image into $M\times N$ grid, and additionally split the vertical space into $K$ levels, as shown in Fig.~\ref{fig:tdp}. In our work, we set $M, N, K$ to 6. 
The annotated images, together with the corresponding sub-task description, are then provided to VLM to generate a sequence of region indices $\mathcal{M}(O_{i,0}, \mathcal{S}_i)= \mathcal{T}_i$, where $\mathcal{T}_i$ represents the trajectory of the sub-task $i$.



\textbf{Voxel-based Spatial Trajectory Representation.} The trajectory generated by the VLM consists of a list of coordinates. We propose representing these coordinates as a $M\times N \times K$ matrix, as illustrated in Fig.~\ref{fig:subtask}. Specifically, the corresponding voxels in the matrix are labeled based on the order of their appearance in the trajectory sequence, while non-trajectory voxels are marked as 0.
This voxel-based spatial trajectory representation facilitates a compact yet spatially informative representation for the downstream diffusion policy. To encode its spatial information, we employ three layers of 3D convolutional neural network (CNN).

\subsection{Trajectory Conditioned Diffusion Policy} \label{sec:diffusion}

Our trajectory-conditioned diffusion policy, similar to the original diffusion policy \cite{chi2023diffusion}, leverages Denoising Diffusion Probabilistic Models (DDPM) \cite{ho2020denoising} to approximate the conditional distribution $p(A_t | O_t, \mathcal{T})$. The process begins by sampling an initial action $A_t^K$ from Gaussian noise. The DDPM then performs $K$ iterations, where at each step, the model predicts the noise level, gradually refining the action and producing an intermediate state $A^{K-1}_{t}$. This iterative process continues until the final, noise-free action $A^0_{t}$ is obtained.
The refinement follows the equation:
$$    
    A^{k-1}_{t} = \alpha (A^k_{t} - \gamma \epsilon_\theta(O_{t}, \mathcal{T}, A^k_t, k) + \mathcal{N}(0, \sigma ^ 2 I)).
$$
In this equation, we introduce a key modification to the original diffusion policy: the noise predictor $\epsilon_\theta$ now accounts not only for the observation $O_t$ and denoising step $k$, but also for the trajectory $\mathcal{T}$. The parameters $\theta$ are now optimized conditioning on this additional trajectory information.

During training, an unmodified action sample $A_t^0$ is drawn along with the corresponding observation $O_t$. A random diffusion step $k$ is selected, and random noise $\epsilon^k$ is added to the action sample. The trajectory conditions are derived from corresponding demonstrations. At each timestep, the position of the robot's end-effector is mapped to the appropriate grid position using the camera’s calibrated extrinsic and intrinsic parameters. We assume that the robot base and the front camera remain fixed throughout the task, as is typical in most manipulation scenarios.
The noise predictor $\epsilon_\theta$ is then trained to predict the added noise $\epsilon^k$ from the noised action sample. The training objective is to minimize the mean squared error (MSE) between the predicted noise and the actual noise. The loss function is defined as: 
$$
    \mathcal{L} = MSE(\epsilon^k, \epsilon_\theta(O_t, \mathcal{T}, A^0_t + \epsilon^k, k)).
$$
In practice, to incorporate trajectory conditioning into the diffusion policy, we introduced several modifications. The encoded trajectory is concatenated with the historical sequence of encoded image observations and the flattened robot state. Importantly, the trajectory information is included only once, regardless of the observation sequence length, under on the assumption that the trajectory remains constant within each sub-task. 

\section{\kefi{Simulation} Experiment Results}

In this section, we report experimental results to address the following pivotal questions: (1) How effective is our trajectory-conditioned approach, compared with other diffusion-based methods? (2) How well does our method, which decomposes complex problems into multiple sub-tasks, enhance the ability to solve long horizon problems? (3) Does the trajectory-conditioned module reduce the reliance on vision, thereby improving system robustness in the presence of various variations?
We extensively evaluate our method on RLBench~\cite{james2020rlbench}. RLBench is built on the CoppelaSim~\cite{rohmer2013v} simulator, where a Franka Panda Robot is used to perform a variety of tasks. \tgli{On RLBench, our model and all baselines were trained to predict next $T$ steps joint positions and execute the first $N$ steps of those $T$. Empirically, we found $T=12$ and $N=8$ work well in our experiments.} We trained and evaluated our model in two setups, differing by tasks and the number of cameras used:

\begin{enumerate} 
    \item \textbf{Single-view setup}, as used in \cite{ze20243d} and \cite{li2024language}. This setup includes a single camera view from the front, as used in the seven RLBench tasks in Section~\ref{exp:ben}.
    \item \textbf{Multi-view setup}, as used in \cite{chi2023diffusion}. This setup uses both the front view and the wrist view of the robot, with the wrist view dynamically moving during task execution, as used in the long horizon tasks in Section~\ref{exp:long}.
\end{enumerate}

\tgli{We separately evaluated the capacity of our Trajectory-conditioned Diffusion Policy (TDP) and VLM-guided Trajectory-conditioned Diffusion Policy (VLM-TDP). TDP takes groundtruth trajectories, aiming to demonstrate the capability of tracking given trajectories. VLM-TDP uses the VLM-generated trajectories, which shows the overall performance.}
We used GPT-4o in our experiments. 
\tgli{To maintain the consistency as in~\cite{li2024language}, we trained policies in RLBench tasks and long horizon tasks for 500 epochs and evaluated them over 20 episodes every 50 epochs. We report the result by averaging the five highest success rates.}

\begin{table*}[ht]
    \caption{\textbf{Evaluation on RLBench Tasks.} Each task was evaluated over 20 episodes, with the average success rate taken across the top 5 checkpoints. Our 2D model, along with the diffusion policy, was compared to 3D policies (3D diffusion and Lang-o3dp). TDP and VLM-TDP outperformed the diffusion policy in all tasks, matching or even exceeding the performance of 3D policies.}
    \vspace{-0.3cm}
    \label{tab:exp_lang_o3dp}
    \centering
    \begin{minipage}{.13\textwidth}
        \centering
        \includegraphics[width=\linewidth]{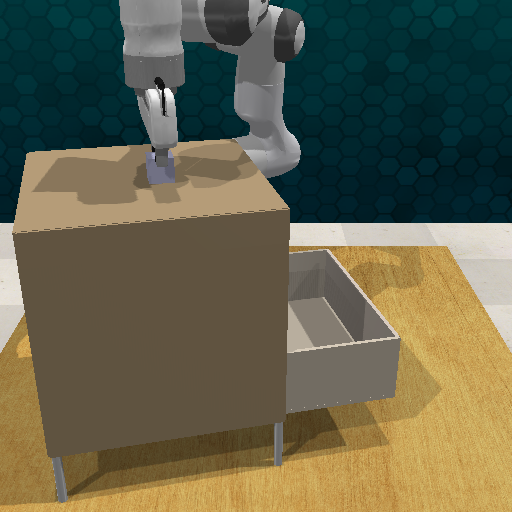}
    \end{minipage}
    \begin{minipage}{.13\textwidth}
        \centering
        \includegraphics[width=\linewidth]{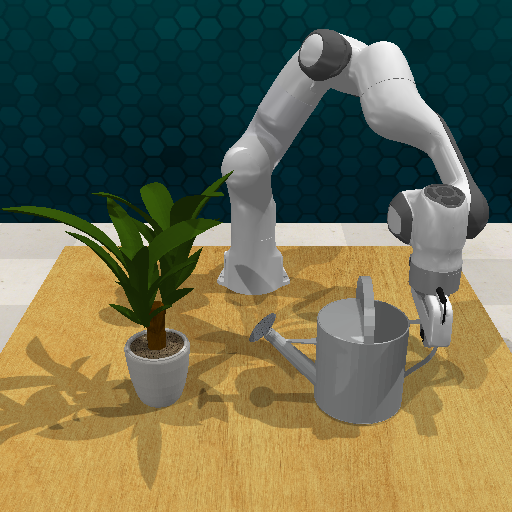} 
    \end{minipage}
    \begin{minipage}{.13\textwidth}
        \centering
        \includegraphics[width=\linewidth]{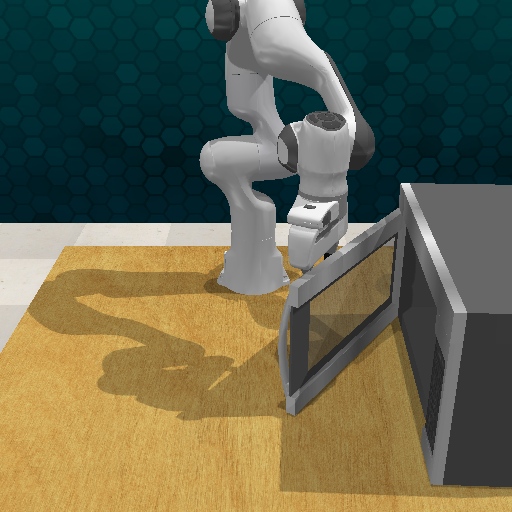}
    \end{minipage}
    \begin{minipage}{.13\textwidth}
        \centering
        \includegraphics[width=\linewidth]{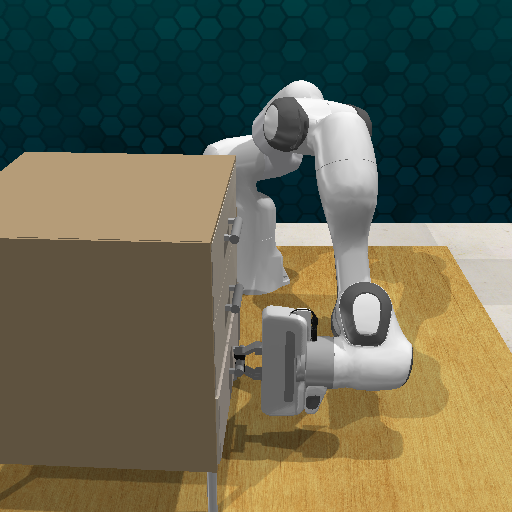}
    \end{minipage}
    \begin{minipage}{.13\textwidth}
       \centering
        \includegraphics[width=\linewidth]{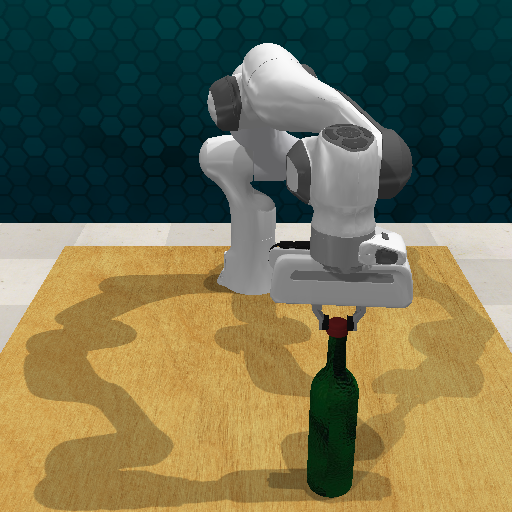}
    \end{minipage}
    \begin{minipage}{.13\textwidth}
        \centering
        \includegraphics[width=\linewidth]{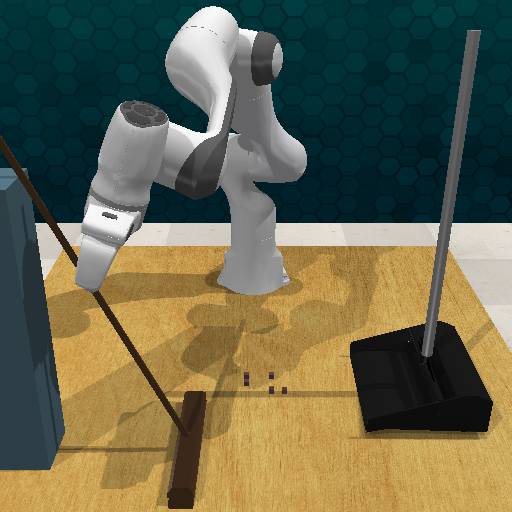}
    \end{minipage}
    \begin{minipage}{.13\textwidth}
        \centering
        \includegraphics[width=\linewidth]{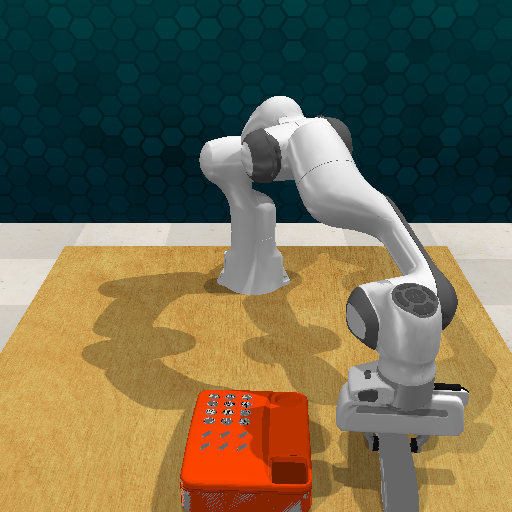}
    \end{minipage}\\
    \vspace{.8em}
    \begin{minipage}[t]{.13\textwidth}
        \centering
        Put Item in Drawer
    \end{minipage}
    \begin{minipage}[t]{.13\textwidth}
        \centering
        Water Plants
    \end{minipage}
    \begin{minipage}[t]{.13\textwidth}
        \centering
        Close Microwave
    \end{minipage}
    \begin{minipage}[t]{.13\textwidth}
        \centering
        Open Drawer
    \end{minipage}
    \begin{minipage}[t]{.13\textwidth}
        \centering
        Open Wine Bottle
    \end{minipage}
    \begin{minipage}[t]{.13\textwidth}
        \centering
        Sweep to Dustpan
    \end{minipage}
    \begin{minipage}[t]{.13\textwidth}
        \centering
        Phone on Base
    \end{minipage}
    \begin{center}
        \renewcommand{\arraystretch}{1.5} 
        \begin{tabular}[b]{
            >{\centering\arraybackslash}m{20ex}
            >{\centering\arraybackslash}m{11ex}
            *8{
                >{\centering\arraybackslash}m{11ex}
            }
    }
            \hline
            & \textbf{Average} & Put Item in Drawer & Water Plants & Close Microwave  & Open Drawer & Open Wine Bottle & Sweep to Dustpan & Phone on Base \\
            \hline
            Diffusion Policy & $0.49$& $0.32$ & $0.41$ & $0.95$  & $0.70$ & $0.38$ & $0.57$ & $0.11$ \\
            TDP (ours) &$\mathbf{0.71}$ & $\mathbf{0.54}$& $\mathbf{0.72}$ &$0.96$ & $0.81$ & $0.62$ & $0.76 $ & $0.55$ \\ 
            VLM-TDP (ours) & $0.69$& $0.52$ & $0.70$ &$\mathbf{0.97}$ & $0.80$ &$0.61$& $0.67$ &$0.53$ \\
            \hline
            3D Diffusion Policy & $0.50$ & $0.05$ & $0.21$ & $0.94$  & $\mathbf{0.94}$ & $0.49$ & $0.66$ & $0.06$ \\
            Lang-o3dp & $0.69$ & $0.50$ & $0.38$ & $0.93$  & $0.90$ & $\mathbf{0.77}$ & $\mathbf{0.77}$ & $\mathbf{0.57}$ \\
            \hline
        \end{tabular}
    \end{center}
    \vspace{-0.5cm}
\end{table*}

\subsection{Evaluation on RLBench Tasks} \label{exp:ben}

We compared our method to three baselines: Diffusion Policy (DP)~\cite{chi2023diffusion}, 3D Diffusion Policy (DP3)~\cite{ze20243d}, and Language-guided object-centric Diffusion Policy (Lang-o3dp)~\cite{li2024language}. Notably, both Diffusion Policy and our method are 2D policies that take RGB images as input, while DP3 and Lang-o3dp are 3D policies that use point cloud and RGB information. 
\tgli{Though it seems unfair to compare 2D policies with 3D policies, given the inherently richer information content in 3D data, such comparative analysis still holds value. The 2D data is often more readily available and has the potential to outperform 3D methods in some scenarios, highlighting the importance of exploring trade-offs and benefits in employing 2D approaches for specific contexts.}
We followed the same setup as~\cite{li2024language} where seven tasks in RLBench were trained and evaluated.
For each task, we generated 40 demonstrations as the training set, and 20 episodes as the testing set. All of the methods follow the single-view setup.

TABLE~\ref{tab:exp_lang_o3dp} presents the results of our experiments across seven tasks. Both TDP and VLM-TDP outperformed diffusion policy in all tasks, demonstrating the advantage of incorporating the trajectory modality into the policy. Despite being a 2D policy, our method achieved the same level of performance in most tasks and even exceed them in specific tasks compared to the state-of-the-art 3D policies, resulting in an improvement of 3\% on average across seven tasks.

In tasks like \emph{Open Drawer}, \emph{Open Wine Bottle}, and \emph{Sweep to Dustpan}, our policy outperformed the diffusion policy but was less effective than 3D policies. In these tasks, the point cloud is less cluttered and the objects are clearly distinguishable. This observation suggests that point cloud can handle distinguishable objects more effectively than RGB information.
However, in tasks such as \emph{Water Plants}, where the plant leaves are crowded or obstructed, 3D policies struggled to accurately evaluate the scene. In contrast, both diffusion policy and our method surpassed 3D policies. These observations show that our method can push the success rate to a higher level for these crowded and cluttered situations.

On the other hand, for tasks like \emph{Phone on Base} and \emph{Put Item on Drawer}, the phone and the item are not easily distinguishable from the table or drawer (both in RGB and point cloud) due to their flat or small shape. Therefore, the success rates of Diffusion Policy and 3D Diffusion Policy were quite low. In contrast, our trajectory-guided policy and Lang-o3dp achieved higher success rates by accurately targeting the object either through trajectory condition or by masking object's point cloud using SAM~\cite{kirillov2023segment}.
Besides, our method excelled in the longer-horizon task \emph{Put Item in Drawer}, which consists of four sub-tasks (grasp handle, open drawer, grasp item, and place it into the drawer). The ability to break down these tasks into sub-tasks allowed our model to benefit from a more structured execution, resulting in superior performance.

\subsection{Evaluation on Long Horizon Tasks} \label{exp:long}

To evaluate our method on solving long horizon problems by decomposing complex problems into sub-tasks, we used a modified version of the \emph{Stack Blocks} in RLBench. The objective is to sequentially pick up all the blocks and stack them at the center. 
Based on our definition of a sub-task, a complete episode typically consists of two types: \emph{Pick} and \emph{Place}. Each sub-task generally requires around 150 timesteps, so stacking four blocks can easily extend beyond 1000 timesteps for the entire episode. We evaluated our model and Diffusion Policy on stacking 1, 2, and 4 blocks. We manually split the task into several sub-tasks and report the results in TABLE~\ref{tab:exp_stack_blocks}.

It can be observed that the sub-task \emph{Place} has higher success rate than the sub-task \emph{Pick} for both models, indicating that \emph{Place} is easier than \emph{Pick}. 
This task involves moving the gripper from its current position while holding a block and placing it at the center.
In contrast, the sub-task \emph{Pick} requires the model to first locate the block and then grasp it. When stacking multiple blocks, the policy also needs to decide the order in which to stack them.
With the help of trajectory conditioning, this decision-making load is reduced for the execution policy, resulting in a 50\% to 200\% increase in performance in \emph{Pick}. This improvement significantly boosts the overall success rate for the complete episode.

When comparing the performance of the policies across different numbers of blocks, both showed a decline in success as the complexity of choices increased. However, the performance drop for our model was smaller. Although the trajectory can theoretically guide the policy in selecting which block to grasp, the generated trajectory can sometimes be ambiguous, especially when two blocks are placed close together. This ambiguity arises from the constraints in planning grid resolution. 
Increasing the grid resolution would make it more challenging for the VLM to generate an accurate trajectory.
Consequently, the success rate drops from 0.88 for picking one block (where there is no ambiguity) to around 0.5 for more complex configurations. 

\begin{table*}[h]
    \caption{\textbf{Evaluation on Long-Horizon Tasks.} Performance comparison across stacking tasks with 1, 2, and 4 blocks. The sub-tasks are categorized into Pick and Place. Comb refers to the evaluation of the overall success of the complete episode.}
    \vspace{-0.3cm}
    \label{tab:exp_stack_blocks}
    \begin{center}
        \renewcommand{\arraystretch}{1.5} 
        \begin{tabular}[b]{
            >{\centering\arraybackslash}m{16ex}
            >{\centering\arraybackslash}m{3ex}
            >{\centering\arraybackslash}m{9ex}
            >{\centering\arraybackslash}m{9ex}|
            >{\centering\arraybackslash}m{9ex}
            >{\centering\arraybackslash}m{3ex}
            >{\centering\arraybackslash}m{9ex}
            >{\centering\arraybackslash}m{9ex}
            >{\centering\arraybackslash}m{9ex}
            >{\centering\arraybackslash}m{9.5ex}|
            >{\centering\arraybackslash}m{9ex}
}
            \hline
            && \multicolumn{3}{c}{Stack \textbf{1} Block} && \multicolumn{5}{c}{Stack \textbf{2} Blocks} \\ 
            \cline{3-5}  \cline{7-11} 
            && Sub-task \textbf{Pick} & Sub-task \textbf{Place} & \textbf{Comb} && Sub-task \textbf{Pick 1st} & Sub-task \textbf{Place 1st} &  Sub-task \textbf{Pick 2nd} & Sub-task \textbf{Place 2nd} & \textbf{Comb} \\
            \hline
            Diffusion Policy && $0.58$ & $0.99$ & $0.26$ && $0.28$  & $\mathbf{0.88}$ & $0.36$ & $0.94$ & $0.02$\\
            TDP (ours) && $\mathbf{0.88}$ & $0.99$ & $0.48$ && $\mathbf{0.58 } $& $0.86$ & $\mathbf{0.68}$ & $\mathbf{0.95}$ &  $\mathbf{0.17} $\\
            VLM-TDP (ours) &&$\mathbf{0.88}$ & $0.99$ & $\mathbf{0.52}$  &  &$0.45$ & $0.85$ & $0.62$& $0.90$ &$0.15$\\
            \hline
        \end{tabular}
    \end{center}
    \begin{center}
        \renewcommand{\arraystretch}{1.5} 
        \begin{tabular}[b]{
            >{\centering\arraybackslash}m{16ex}
            >{\centering\arraybackslash}m{9ex}
            >{\centering\arraybackslash}m{9ex}
            >{\centering\arraybackslash}m{9ex}
            >{\centering\arraybackslash}m{9.5ex}
            >{\centering\arraybackslash}m{9ex}
            >{\centering\arraybackslash}m{9.5ex}
            >{\centering\arraybackslash}m{9ex}
            >{\centering\arraybackslash}m{9ex}|
            >{\centering\arraybackslash}m{9ex}
}
            \hline
            & \multicolumn{9}{c}{Stack \textbf{4} Blocks} \\ 
            \cline{2-10} 
            & Sub-task \textbf{Pick 1st} & Sub-task \textbf{Place 1st} & Sub-task \textbf{Pick 2nd} &Sub-task \textbf{Place 2nd} & Sub-task \textbf{Pick 3rd} & Sub-task \textbf{Place 3rd} &  Sub-task \textbf{Pick 4th} & Sub-task \textbf{Place 4th} & \textbf{Comb} \\
            \hline
            Diffusion Policy & $0.20$ & $\mathbf{0.82}$ & $0.15$ & $\mathbf{0.80}$ & $0.20$  & $0.69$ & $0.24$ & $\mathbf{0.68}$ & $0.00$\\
            TDP (ours) & $\mathbf{0.47}$ & $0.78$ & $0.44$ & $0.76$ & $\mathbf{0.46}$ & $\mathbf{0.77}$ & $0.30$ & $0.56$ &  $\mathbf{0.05} $\\
            VLM-TDP (ours) & $0.46$ & $0.79$ & $\mathbf{0.47}$ & $0.75$ & $0.44$ & $0.73$& $\mathbf{0.33}$&  $0.55$ & $0.04$\\
            \hline
        \end{tabular}
    \end{center}
\end{table*}

\subsection{Robustness Analysis on Variations} \label{sec:robustness}

\begin{figure}
    \centering
    \includegraphics[width=0.32\linewidth]{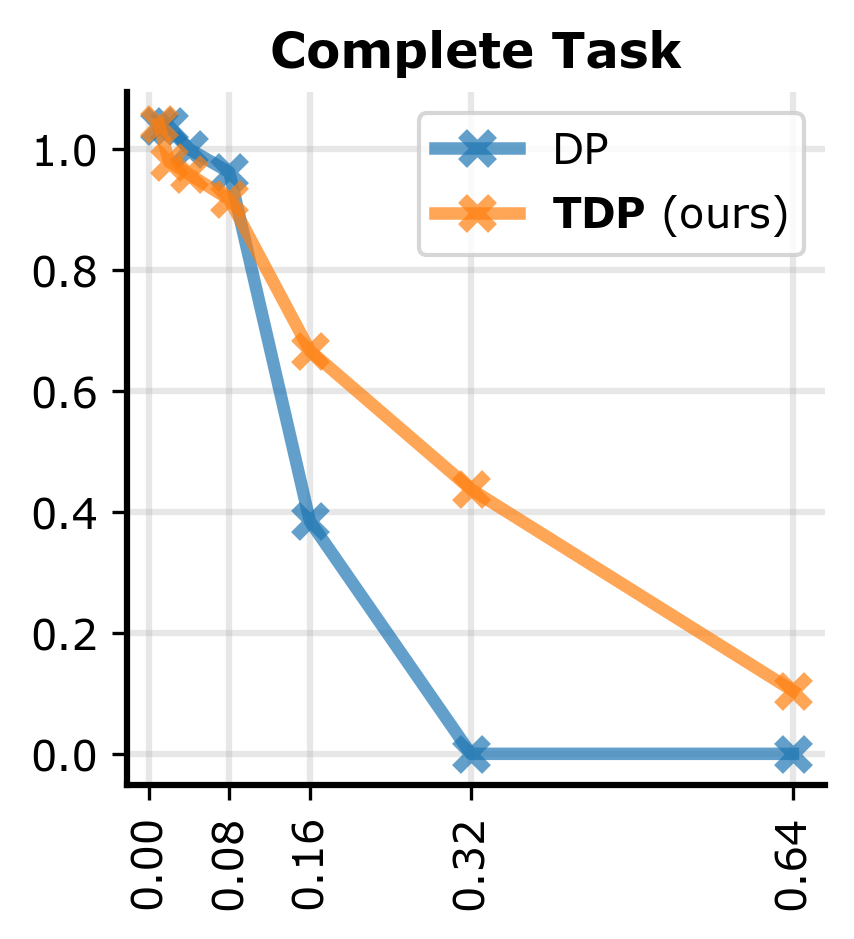}
    \includegraphics[width=0.32\linewidth]{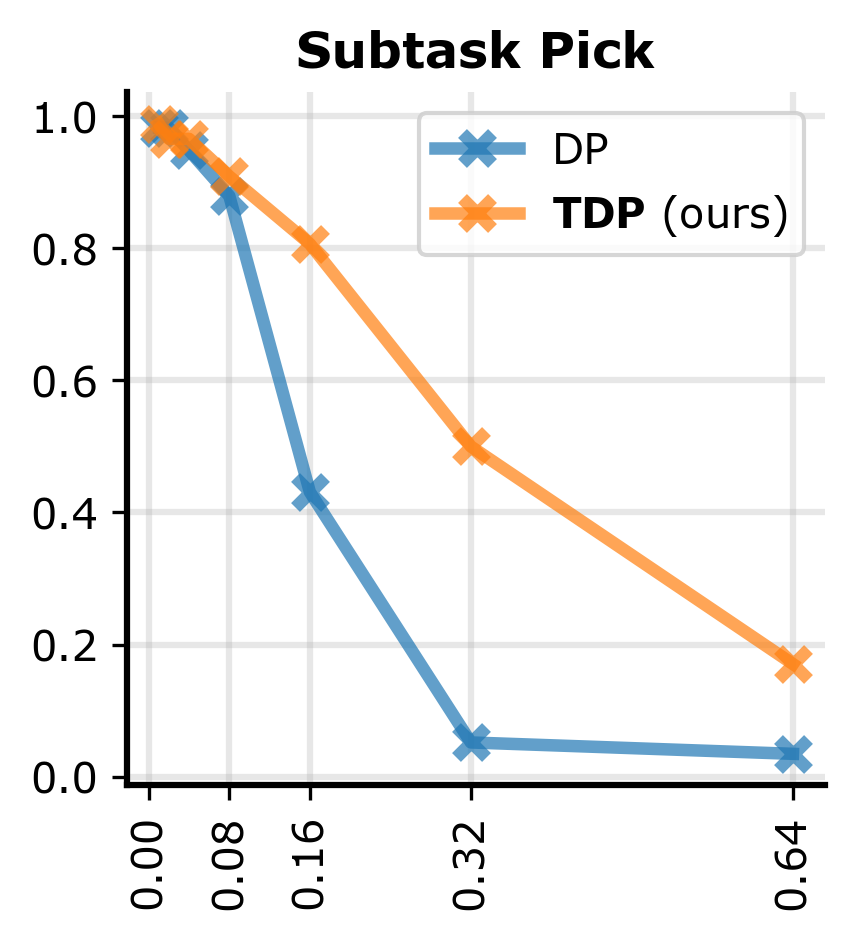}
    \includegraphics[width=0.32\linewidth]{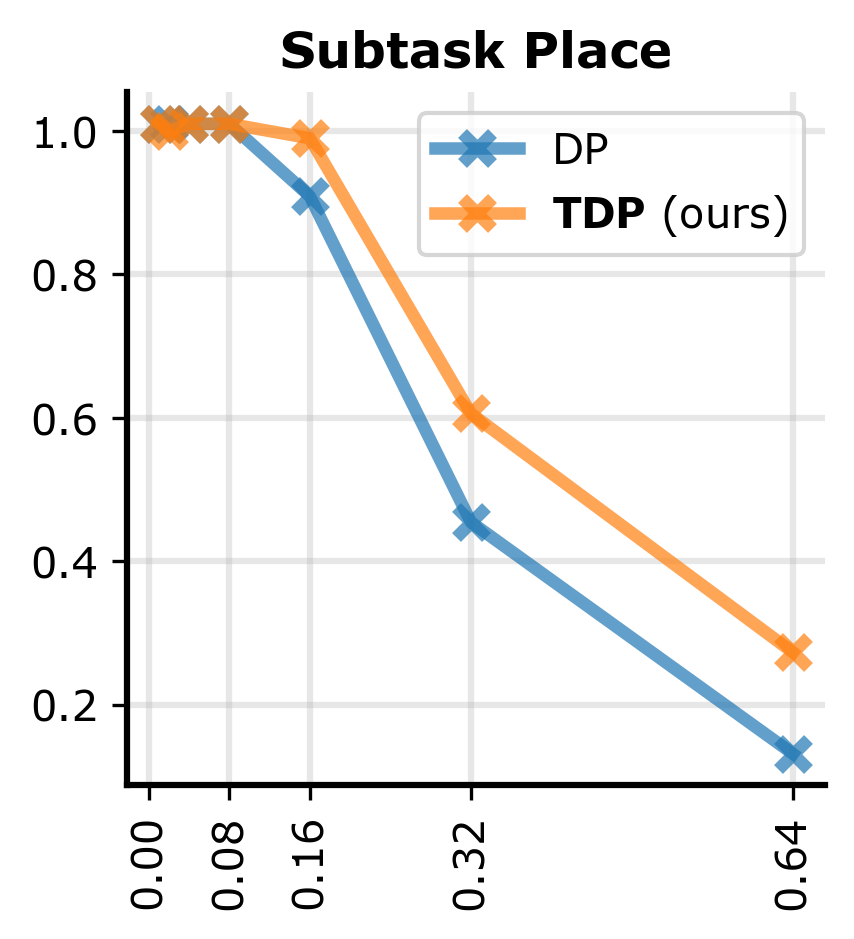}\\
    \vspace{0.5em}
    \begin{minipage}{.95\linewidth}
        \centering
        \includegraphics[width=0.24\linewidth]{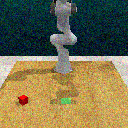}
        \includegraphics[width=0.24\linewidth]{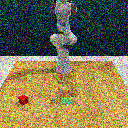}
        \includegraphics[width=0.24\linewidth]{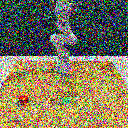}
        \includegraphics[width=0.24\linewidth]{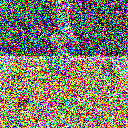}
    \end{minipage}\\
    \vspace{0.5em}
    \begin{minipage}{0.24\linewidth}
        \centering
        $\sigma = 0.08$
    \end{minipage}
    \hfill
    \begin{minipage}{0.23\linewidth}
        \centering
        $\sigma = 0.16$
    \end{minipage}
    \hfill
    \begin{minipage}{0.23\linewidth}
        \centering
        $\sigma = 0.32$
    \end{minipage}
    \hfill
    \begin{minipage}{0.24\linewidth}
        \centering
        $\sigma = 0.64$
    \end{minipage}
    \label{fig:noise_example}
    \caption{\textbf{Relative Success Rate with Noisy Image Input.} Both our model and the diffusion policy were trained on clean images and evaluated using noisy front and wrist image inputs with varying standard deviations ($\sigma$). The top figure presents the success rate relative to the result obtained with clean input. Example images of different noise level are shown at the bottom.}
    \label{fig:exp_noisy_input}
    \vspace{-0.5cm}
\end{figure}

In this section, we evaluate the robustness under different variations and compare it with the Diffusion Policy in the task \emph{Stack 1 Block}. We test the policies under different variations without additional training. 
We first tested it under noisy image inputs by manually adding Gaussian noise with varying standard deviations. We report the relative success rate, compared to the clean input results, both sub-tasks and the complete episode. The results are presented in Fig.~\ref{fig:exp_noisy_input}.

Our model demonstrated greater robustness across all sub-tasks and the complete episode compared to the Diffusion Policy. In the simpler \emph{Place} sub-task, where the robot moves to a fixed target location, the image input plays a less critical role, and low-level robot internal data is sufficient for planning actions. As a result, both models maintained a reasonable success rate even with significant noise ($\sigma = 0.64$), although our model exhibited a smaller performance drop due to the added trajectory condition.
In contrast, the \emph{Pick} sub-task is more challenging, as the policy requires accurate information about the block's location. For the diffusion policy, the image is the sole source of information, and its performance deteriorated sharply under noise, reaching a $0\%$ success rate at $\sigma = 0.32$. However, our model benefits from the trajectory guidance, which helps direct the end-effector closer to the block, allowing the wrist camera to compensate for the noise at shorter distances. Consequently, TDP maintained $17.0\%$ of its original performance, with an absolute success rate of $0.15$ at $\sigma = 0.64$ in \emph{Pick}, and achieved an $10.4\%$ of its original complete task performance.

Next, we introduced more variants to the environment by leveraging the recent work Colosseum~\cite{pumacay2024colosseum}. We selected four variations: background texture, object texture, smaller objects and larger objects. We show the results in Fig.4.
Our policy consistently exhibited a smaller performance drop compared to the diffusion policy across all variations. Notably, when increasing the object size, both policies showed improved performance in the \emph{Pick} sub-task but decreased performance in the \emph{Place} sub-task. Interestingly, our model also demonstrated an overall improvement in performance with larger object sizes. 
We hypothesize that as the object size increases, the precision required for the robot’s execution decreases, reducing the difficulty for the backbone execution policy. As a result, our trajectory conditioning plays a more significant role in enhancing performance.

\begin{figure}     \label{fig:exp_colosseum}
    \centering
    \vspace{-0.5cm}
    \includegraphics[width=.49\linewidth]{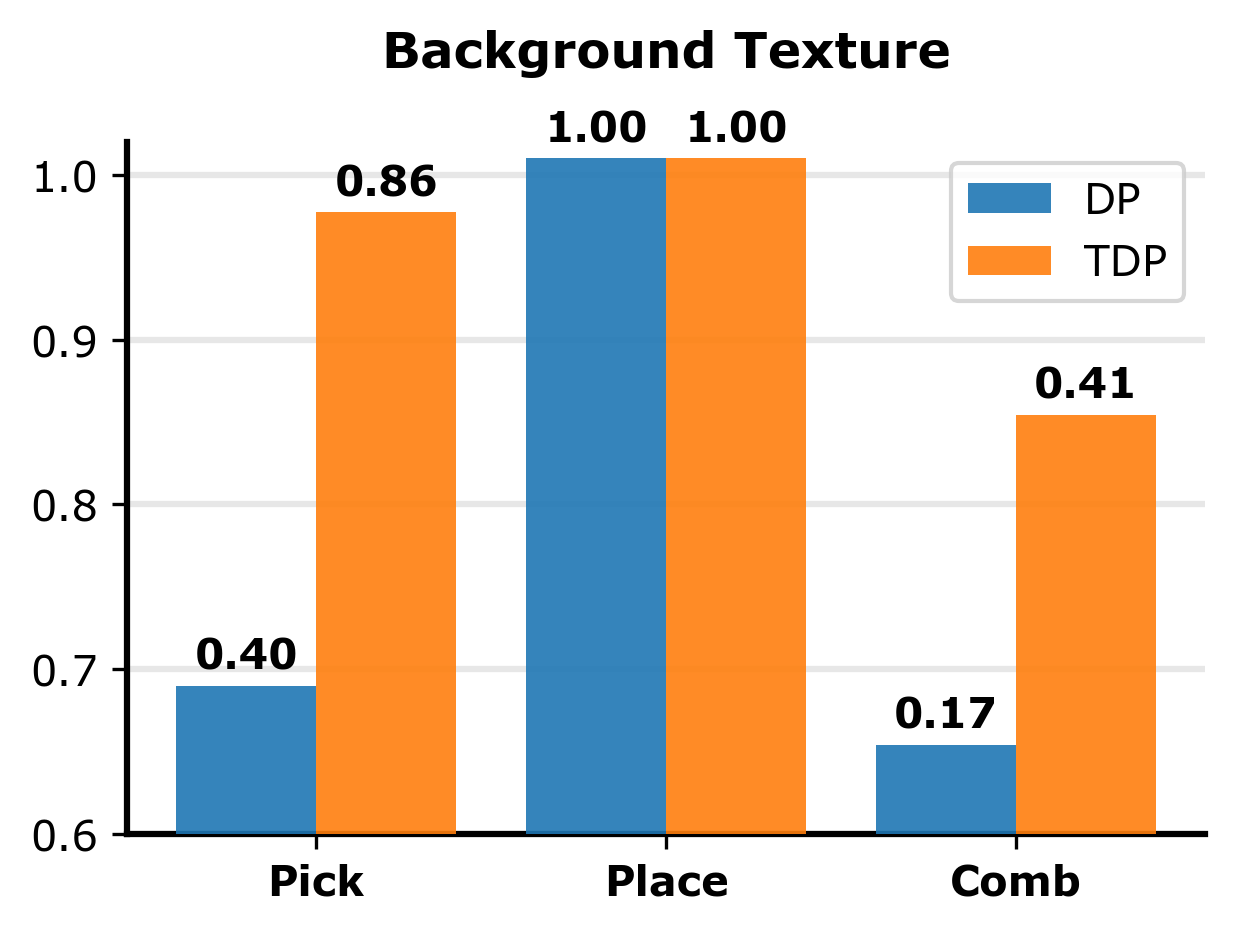}
    \includegraphics[width=.49\linewidth]{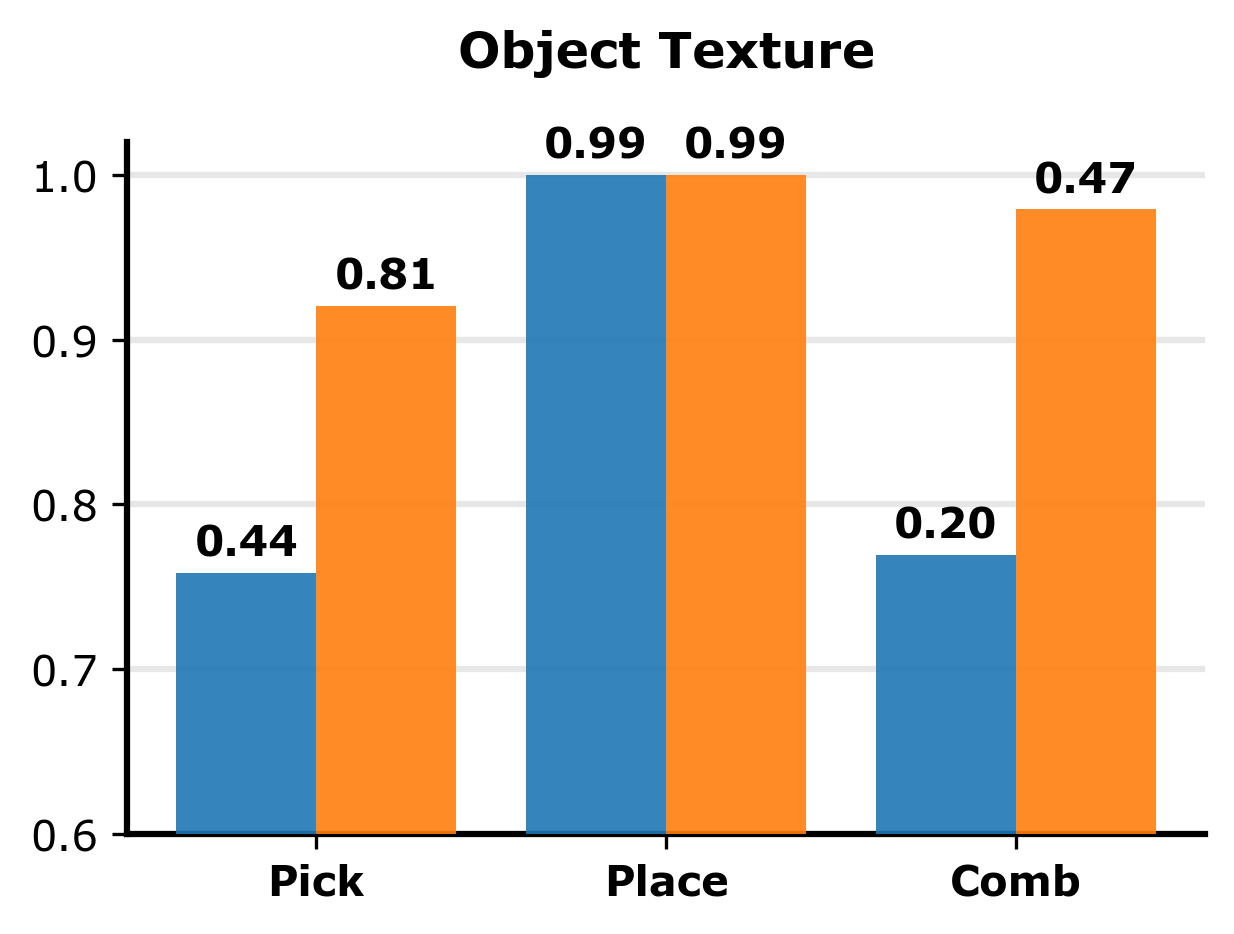}
    \includegraphics[width=.49\linewidth]{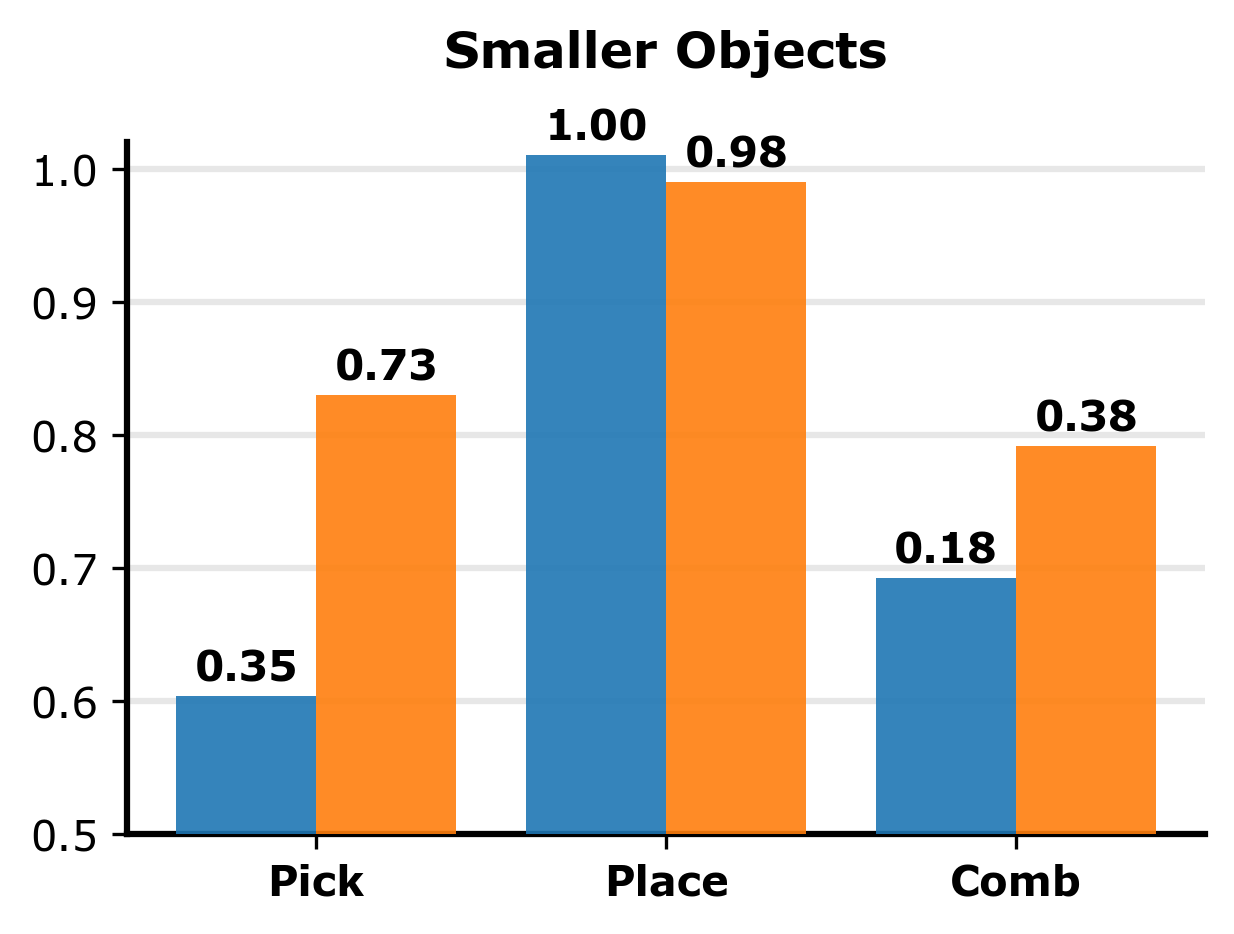}
    \includegraphics[width=.49\linewidth]{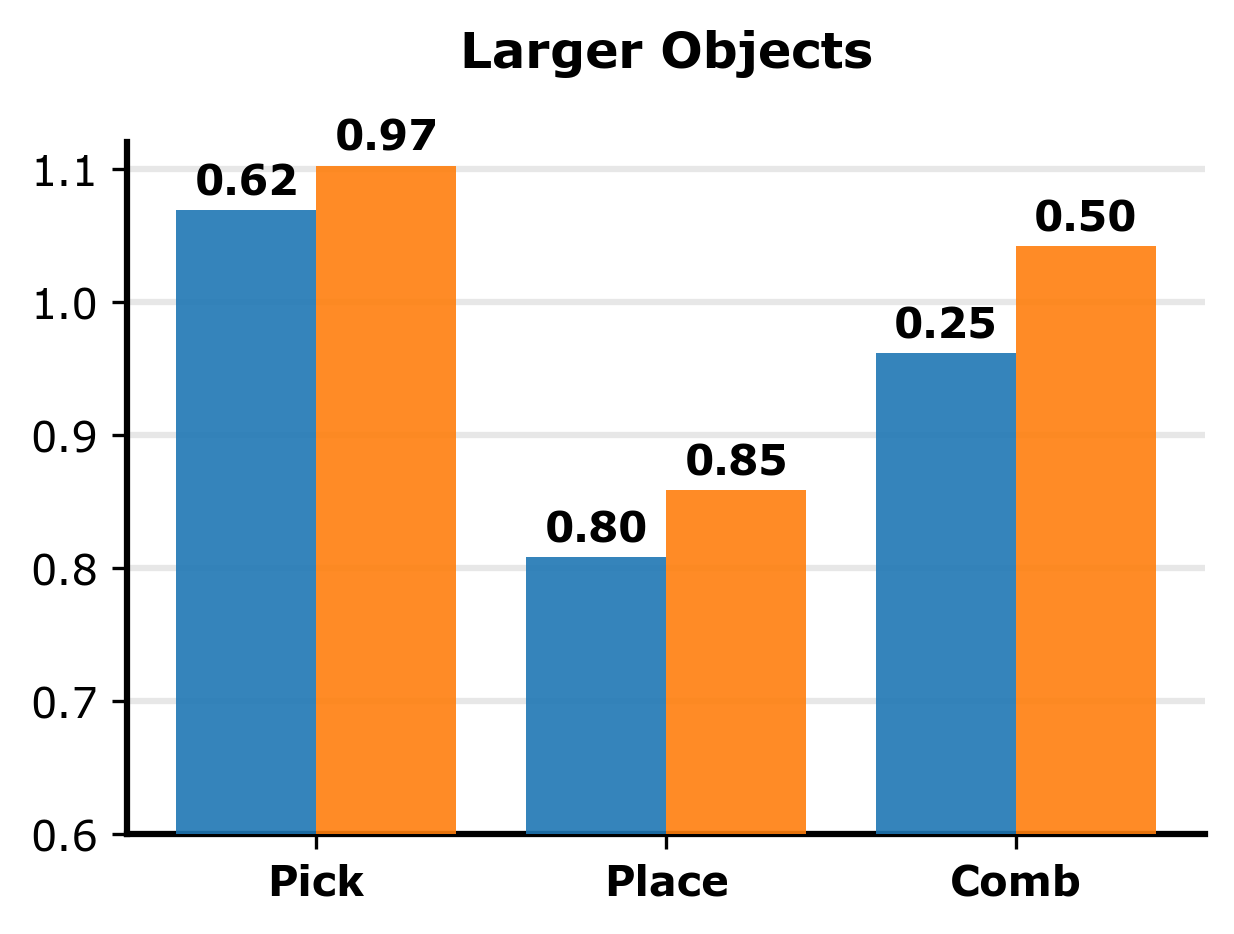}\\
    \vspace{0.5em}
    \begin{minipage}{0.95\linewidth}
        \includegraphics[width=0.24\linewidth]{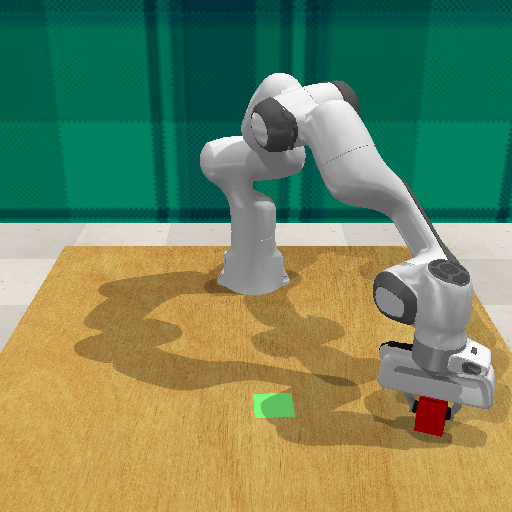}
        \includegraphics[width=0.24\linewidth]{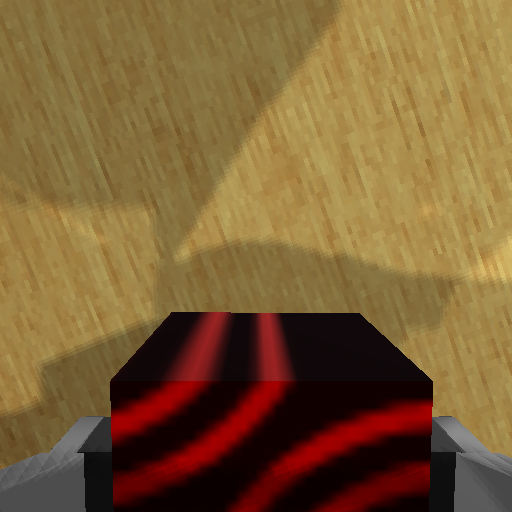}
        \includegraphics[width=0.24\linewidth]{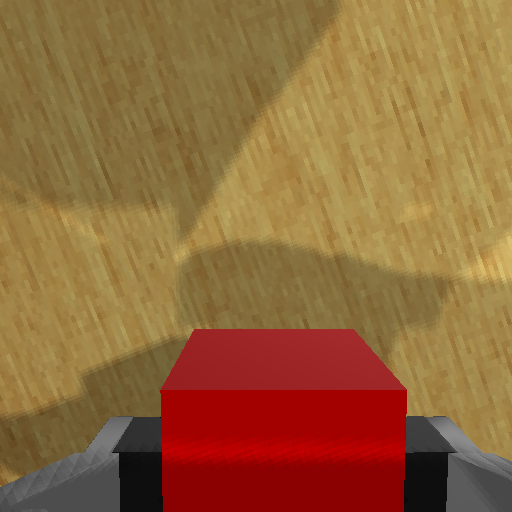}
        \includegraphics[width=0.24\linewidth]{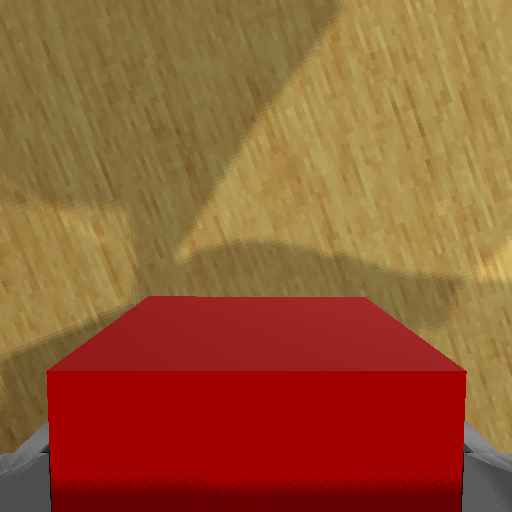}
    \end{minipage}\\ 
    \vspace{0.5em}
    \begin{minipage}{0.95\linewidth}
        \begin{minipage}[t]{0.24\linewidth}
            \centering
            Background Texture
        \end{minipage}
        \begin{minipage}[t]{0.24\linewidth}
            \centering
            Object Texture
        \end{minipage}
        \begin{minipage}[t]{0.24\linewidth}
            \centering
            Smaller Object
        \end{minipage}
        \begin{minipage}[t]{0.24\linewidth}
            \centering
            Larger \\Object
        \end{minipage}
    \end{minipage}\\
    \caption{\textbf{Success Rates with Colosseum Variations.} The top figures present a comparison of the relative success rates of our TDP model and the diffusion policy (DP) across various task variations. The relative success rate is calculated with respect to the performance under no variations. The absolute success rates are annotated above each bar, and example images of the variations are displayed below. Our model demonstrates superior generalization across all variations compared to the diffusion policy}
\end{figure}

\section{Real-World Experiment}
\begin{figure}[t]
    \centering
    \includegraphics[width=0.7\linewidth]{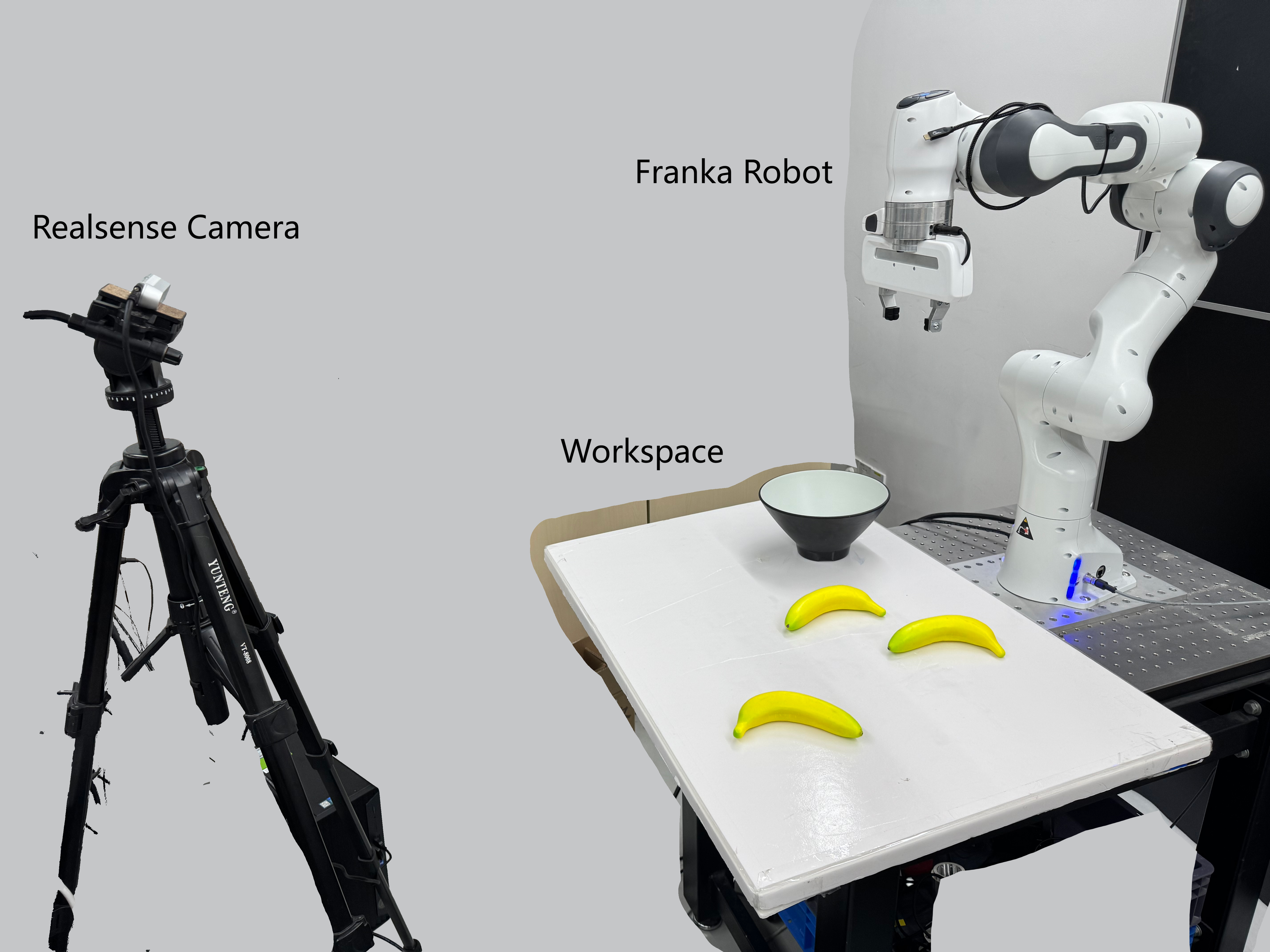}
    \caption{\textbf{Real-world experiment setup.} We use a Franka robotic arm equipped with a parallel gripper for all manipulation tasks. Visual observations are captured using an Intel RealSense camera.}
    \label{fig:robot_setup}
    \vspace{-0.5cm}
\end{figure}

\begin{figure}[t]
    \centering
    \includegraphics[width=0.7\linewidth]{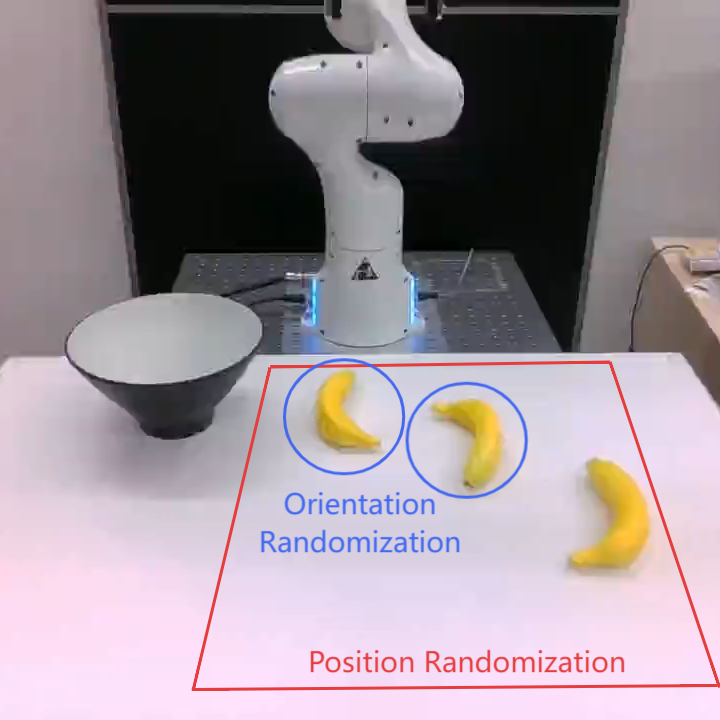}
    \caption{\textbf{Task Randomization.} Object positions are randomized across all trials. For banana tasks, object orientations are also randomized to increase variability and challenge policy robustness.}
    \label{fig:real_random}
\end{figure}
\subsection{Experiment Setup}

\textbf{Robot Setup}. We evaluate all algorithms on the Franka Emika Panda robot equipped with a parallel gripper. Visual observations are captured using a RealSense D435i camera. 

\textbf{Task Definition}. 
We evaluated our method on three tasks: \emph{Pick One Orange}, \emph{Pick Two Bananas}, \emph{Pick Three Bananas}. In each task, the robot aims to sequentially pick up the corresponding objects and place them into a bowl as shown in Fig.~\ref{fig:robot_setup}. These tasks are designed to evaluate algorithm performance in long-horizon manipulation scenarios. We chose an orange for the single-object task because its round shape makes it harder to grasp reliably compared to a banana. The gripper must be precisely positioned to avoid slipping or excessive pressure. Fig.~\ref{fig:real_random} illustrates the randomized conditions for the tasks. All objects are randomized in position and orientation across trials.


\textbf{Data Collection.} Expert demonstrations were collected via human teleoperation using a 6-DOF SpaceMouse device. For each task, we collected 40 successful episodes. 

\textbf{Evaluation Method.} Each task is evaluated across 5 unique randomized setups, with 4 trials per setup, yielding 20 evaluations per task. All setups were manually verified to ensure consistent initial conditions, minimizing the impact of stochastic variations.

\subsection{Experiment Result}
The results are presented in Table~\ref{tab:real_world}. Consistent with our simulation findings, our method outperforms the baseline—particularly in long-horizon tasks. These findings validate the practical applicability and real-world effectiveness of our algorithm on physical robotic platforms.

As task complexity increases, the performance gap becomes more pronounced. For the two-banana task, our method achieves a 95\% success rate (19/20), compared to the baseline Diffusion Policy’s 70\%. In the more demanding three-banana task, our method maintains a solid 70\% success rate, while the baseline drops significantly to just 20\%. Even in the challenging one-orange task—requiring precise and careful grasping—our method demonstrates a clear advantage. This is largely attributed to the guidance provided by the VLM trajectory and the policy’s targeted focus on object grasping.

\begin{table}[t]
    \caption{\kefi{\textbf{Real-World Experiment Results.} Our method consistently outperforms the baseline diffusion policy across tasks of increasing complexity. }}
    \vspace{-0.3cm}
    \label{tab:real_world}
    \centering
    \begin{center}
        \renewcommand{\arraystretch}{1.5} 
        \begin{tabular}[b]{
            >{\centering\arraybackslash}m{20ex}
            >{\centering\arraybackslash}m{11ex}
            *8{
                >{\centering\arraybackslash}m{11ex}
            }
    }
            \hline
            & One Orange & Two Bananas & Three Bananas \\
            \hline
            Diffusion Policy & $0.70$& $0.70$ & $0.20$ \\
            TDP (ours) &$\mathbf{0.85}$ & $\mathbf{0.95}$& $\mathbf{0.70}$ \\ 
            \hline
        \end{tabular}
    \end{center}
    \vspace{-0.5cm}
\end{table}

\section{Conclusion}
In this work, we propose the VLM-TDP method that leverages VLMs to decompose complex tasks into multiple concise sub-tasks, and generate voxel-based spatial trajectories for each sub-task. These trajectories are then used to guide the proposed trajectory-conditioned diffusion policy for action generation.
Experimental results reveal that the proposed method outperforms the diffusion strategy in all test cases. In long-horizon tasks, our approach demonstrates an improvement of over 100\%, highlighting the beneficial effects of sub-tasks configurations. In the presene of varying input images, our method also performs significantly better, indicating that the proposed voxel-based trajectory modalities can effectively enhance the diffusion strategy to generate actions.




\bibliographystyle{IEEEtran}
\bibliography{ref}

\begin{thebibliography}{10}
\providecommand{\url}[1]{#1}
\csname url@samestyle\endcsname
\providecommand{\newblock}{\relax}
\providecommand{\bibinfo}[2]{#2}
\providecommand{\BIBentrySTDinterwordspacing}{\spaceskip=0pt\relax}
\providecommand{\BIBentryALTinterwordstretchfactor}{4}
\providecommand{\BIBentryALTinterwordspacing}{\spaceskip=\fontdimen2\font plus
\BIBentryALTinterwordstretchfactor\fontdimen3\font minus \fontdimen4\font\relax}
\providecommand{\BIBforeignlanguage}[2]{{%
\expandafter\ifx\csname l@#1\endcsname\relax
\typeout{** WARNING: IEEEtran.bst: No hyphenation pattern has been}%
\typeout{** loaded for the language `#1'. Using the pattern for}%
\typeout{** the default language instead.}%
\else
\language=\csname l@#1\endcsname
\fi
#2}}
\providecommand{\BIBdecl}{\relax}
\BIBdecl

\bibitem{mandlekar2021matters}
A.~Mandlekar, D.~Xu, J.~Wong, S.~Nasiriany, C.~Wang, R.~Kulkarni, L.~Fei-Fei, S.~Savarese, Y.~Zhu, and R.~Mart{\'\i}n-Mart{\'\i}n, ``What matters in learning from offline human demonstrations for robot manipulation,'' \emph{arXiv preprint arXiv:2108.03298}, 2021.

\bibitem{cui2022play}
Z.~J. Cui, Y.~Wang, N.~M.~M. Shafiullah, and L.~Pinto, ``From play to policy: Conditional behavior generation from uncurated robot data,'' \emph{arXiv preprint arXiv:2210.10047}, 2022.

\bibitem{florence2019self}
P.~Florence, L.~Manuelli, and R.~Tedrake, ``Self-supervised correspondence in visuomotor policy learning,'' \emph{IEEE Robotics and Automation Letters}, vol.~5, no.~2, pp. 492--499, 2019.

\bibitem{chi2023diffusion}
C.~Chi, S.~Feng, Y.~Du, Z.~Xu, E.~Cousineau, B.~Burchfiel, and S.~Song, ``Diffusion policy: Visuomotor policy learning via action diffusion,'' \emph{arXiv preprint arXiv:2303.04137}, 2023.

\bibitem{ze20243d}
Y.~Ze, G.~Zhang, K.~Zhang, C.~Hu, M.~Wang, and H.~Xu, ``3d diffusion policy: Generalizable visuomotor policy learning via simple 3d representations,'' in \emph{ICRA 2024 Workshop on 3D Visual Representations for Robot Manipulation}, 2024.

\bibitem{ho2020denoising}
J.~Ho, A.~Jain, and P.~Abbeel, ``Denoising diffusion probabilistic models,'' \emph{Advances in neural information processing systems}, vol.~33, pp. 6840--6851, 2020.

\bibitem{ke20243d}
T.-W. Ke, N.~Gkanatsios, and K.~Fragkiadaki, ``3d diffuser actor: Policy diffusion with 3d scene representations,'' \emph{arXiv preprint arXiv:2402.10885}, 2024.

\bibitem{li2024language}
H.~Li, Q.~Feng, Z.~Zheng, J.~Feng, and A.~Knoll, ``Language-guided object-centric diffusion policy for collision-aware robotic manipulation,'' \emph{arXiv preprint arXiv:2407.00451}, 2024.

\bibitem{achiam2023gpt}
J.~Achiam, S.~Adler, S.~Agarwal, L.~Ahmad, I.~Akkaya, F.~L. Aleman, D.~Almeida, J.~Altenschmidt, S.~Altman, S.~Anadkat \emph{et~al.}, ``Gpt-4 technical report,'' \emph{arXiv preprint arXiv:2303.08774}, 2023.

\bibitem{radford2021learning}
A.~Radford, J.~W. Kim, C.~Hallacy, A.~Ramesh, G.~Goh, S.~Agarwal, G.~Sastry, A.~Askell, P.~Mishkin, J.~Clark \emph{et~al.}, ``Learning transferable visual models from natural language supervision,'' in \emph{International conference on machine learning}.\hskip 1em plus 0.5em minus 0.4em\relax PMLR, 2021, pp. 8748--8763.

\bibitem{chowdhery2023palm}
A.~Chowdhery, S.~Narang, J.~Devlin, M.~Bosma, G.~Mishra, A.~Roberts, P.~Barham, H.~W. Chung, C.~Sutton, S.~Gehrmann \emph{et~al.}, ``Palm: Scaling language modeling with pathways,'' \emph{Journal of Machine Learning Research}, vol.~24, no. 240, pp. 1--113, 2023.

\bibitem{ahn2022can}
M.~Ahn, A.~Brohan, N.~Brown, Y.~Chebotar, O.~Cortes, B.~David, C.~Finn, C.~Fu, K.~Gopalakrishnan, K.~Hausman \emph{et~al.}, ``Do as i can, not as i say: Grounding language in robotic affordances,'' \emph{arXiv preprint arXiv:2204.01691}, 2022.

\bibitem{chen2023open}
B.~Chen, F.~Xia, B.~Ichter, K.~Rao, K.~Gopalakrishnan, M.~S. Ryoo, A.~Stone, and D.~Kappler, ``Open-vocabulary queryable scene representations for real world planning,'' in \emph{2023 IEEE International Conference on Robotics and Automation (ICRA)}.\hskip 1em plus 0.5em minus 0.4em\relax IEEE, 2023, pp. 11\,509--11\,522.

\bibitem{huang2022language}
W.~Huang, P.~Abbeel, D.~Pathak, and I.~Mordatch, ``Language models as zero-shot planners: Extracting actionable knowledge for embodied agents,'' in \emph{International conference on machine learning}.\hskip 1em plus 0.5em minus 0.4em\relax PMLR, 2022, pp. 9118--9147.

\bibitem{chen2024spatialvlm}
B.~Chen, Z.~Xu, S.~Kirmani, B.~Ichter, D.~Sadigh, L.~Guibas, and F.~Xia, ``Spatialvlm: Endowing vision-language models with spatial reasoning capabilities,'' in \emph{Proceedings of the IEEE/CVF Conference on Computer Vision and Pattern Recognition}, 2024, pp. 14\,455--14\,465.

\bibitem{yang2023set}
J.~Yang, H.~Zhang, F.~Li, X.~Zou, C.~Li, and J.~Gao, ``Set-of-mark prompting unleashes extraordinary visual grounding in gpt-4v,'' \emph{arXiv preprint arXiv:2310.11441}, 2023.

\bibitem{liang2023code}
J.~Liang, W.~Huang, F.~Xia, P.~Xu, K.~Hausman, B.~Ichter, P.~Florence, and A.~Zeng, ``Code as policies: Language model programs for embodied control,'' in \emph{2023 IEEE International Conference on Robotics and Automation (ICRA)}.\hskip 1em plus 0.5em minus 0.4em\relax IEEE, 2023, pp. 9493--9500.

\bibitem{huang2023voxposer}
W.~Huang, C.~Wang, R.~Zhang, Y.~Li, J.~Wu, and L.~Fei-Fei, ``Voxposer: Composable 3d value maps for robotic manipulation with language models,'' \emph{arXiv preprint arXiv:2307.05973}, 2023.

\bibitem{liu2024moka}
F.~Liu, K.~Fang, P.~Abbeel, and S.~Levine, ``Moka: Open-vocabulary robotic manipulation through mark-based visual prompting,'' \emph{arXiv preprint arXiv:2403.03174}, 2024.

\bibitem{james2020rlbench}
S.~James, Z.~Ma, D.~R. Arrojo, and A.~J. Davison, ``Rlbench: The robot learning benchmark \& learning environment,'' \emph{IEEE Robotics and Automation Letters}, vol.~5, no.~2, pp. 3019--3026, 2020.

\bibitem{pumacay2024colosseum}
W.~Pumacay, I.~Singh, J.~Duan, R.~Krishna, J.~Thomason, and D.~Fox, ``The colosseum: A benchmark for evaluating generalization for robotic manipulation,'' \emph{arXiv preprint arXiv:2402.08191}, 2024.

\bibitem{sohl2015deep}
J.~Sohl-Dickstein, E.~Weiss, N.~Maheswaranathan, and S.~Ganguli, ``Deep unsupervised learning using nonequilibrium thermodynamics,'' in \emph{International conference on machine learning}.\hskip 1em plus 0.5em minus 0.4em\relax PMLR, 2015, pp. 2256--2265.

\bibitem{song2020denoising}
J.~Song, C.~Meng, and S.~Ermon, ``Denoising diffusion implicit models,'' \emph{arXiv preprint arXiv:2010.02502}, 2020.

\bibitem{song2020score}
Y.~Song, J.~Sohl-Dickstein, D.~P. Kingma, A.~Kumar, S.~Ermon, and B.~Poole, ``Score-based generative modeling through stochastic differential equations,'' \emph{arXiv preprint arXiv:2011.13456}, 2020.

\bibitem{rombach2022high}
R.~Rombach, A.~Blattmann, D.~Lorenz, P.~Esser, and B.~Ommer, ``High-resolution image synthesis with latent diffusion models,'' in \emph{Proceedings of the IEEE/CVF conference on computer vision and pattern recognition}, 2022, pp. 10\,684--10\,695.

\bibitem{yu2023scaling}
T.~Yu, T.~Xiao, A.~Stone, J.~Tompson, A.~Brohan, S.~Wang, J.~Singh, C.~Tan, J.~Peralta, B.~Ichter \emph{et~al.}, ``Scaling robot learning with semantically imagined experience,'' \emph{arXiv preprint arXiv:2302.11550}, 2023.

\bibitem{mandi2022cacti}
Z.~Mandi, H.~Bharadhwaj, V.~Moens, S.~Song, A.~Rajeswaran, and V.~Kumar, ``Cacti: A framework for scalable multi-task multi-scene visual imitation learning,'' \emph{arXiv preprint arXiv:2212.05711}, 2022.

\bibitem{chen2023genaug}
Z.~Chen, S.~Kiami, A.~Gupta, and V.~Kumar, ``Genaug: Retargeting behaviors to unseen situations via generative augmentation,'' \emph{arXiv preprint arXiv:2302.06671}, 2023.

\bibitem{chen2024dreamarrangement}
W.~Chen, C.~Xiao, G.~Gao, F.~Sun, C.~Zhang, and J.~Zhang, ``Dreamarrangement: Learning language-conditioned robotic rearrangement of objects via denoising diffusion and vlm planner,'' \emph{Authorea Preprints}, 2024.

\bibitem{bartsch2024sculptdiff}
A.~Bartsch, A.~Car, C.~Avra, and A.~B. Farimani, ``Sculptdiff: Learning robotic clay sculpting from humans with goal conditioned diffusion policy,'' \emph{arXiv preprint arXiv:2403.10401}, 2024.

\bibitem{ma2024hierarchical}
X.~Ma, S.~Patidar, I.~Haughton, and S.~James, ``Hierarchical diffusion policy for kinematics-aware multi-task robotic manipulation,'' in \emph{Proceedings of the IEEE/CVF Conference on Computer Vision and Pattern Recognition}, 2024, pp. 18\,081--18\,090.

\bibitem{bousmalis2023robocat}
K.~Bousmalis, G.~Vezzani, D.~Rao, C.~Devin, A.~X. Lee, M.~Bauza, T.~Davchev, Y.~Zhou, A.~Gupta, A.~Raju \emph{et~al.}, ``Robocat: A self-improving foundation agent for robotic manipulation,'' \emph{arXiv preprint arXiv:2306.11706}, 2023.

\bibitem{lynch2020learning}
C.~Lynch, M.~Khansari, T.~Xiao, V.~Kumar, J.~Tompson, S.~Levine, and P.~Sermanet, ``Learning latent plans from play,'' in \emph{Conference on robot learning}.\hskip 1em plus 0.5em minus 0.4em\relax PMLR, 2020, pp. 1113--1132.

\bibitem{jang2022bc}
E.~Jang, A.~Irpan, M.~Khansari, D.~Kappler, F.~Ebert, C.~Lynch, S.~Levine, and C.~Finn, ``Bc-z: Zero-shot task generalization with robotic imitation learning,'' in \emph{Conference on Robot Learning}.\hskip 1em plus 0.5em minus 0.4em\relax PMLR, 2022, pp. 991--1002.

\bibitem{brohan2023rt}
A.~Brohan, N.~Brown, J.~Carbajal, Y.~Chebotar, X.~Chen, K.~Choromanski, T.~Ding, D.~Driess, A.~Dubey, C.~Finn \emph{et~al.}, ``Rt-2: Vision-language-action models transfer web knowledge to robotic control,'' \emph{arXiv preprint arXiv:2307.15818}, 2023.

\bibitem{nair2022learning}
S.~Nair, E.~Mitchell, K.~Chen, S.~Savarese, C.~Finn \emph{et~al.}, ``Learning language-conditioned robot behavior from offline data and crowd-sourced annotation,'' in \emph{Conference on Robot Learning}.\hskip 1em plus 0.5em minus 0.4em\relax PMLR, 2022, pp. 1303--1315.

\bibitem{stone2023open}
A.~Stone, T.~Xiao, Y.~Lu, K.~Gopalakrishnan, K.-H. Lee, Q.~Vuong, P.~Wohlhart, S.~Kirmani, B.~Zitkovich, F.~Xia \emph{et~al.}, ``Open-world object manipulation using pre-trained vision-language models,'' \emph{arXiv preprint arXiv:2303.00905}, 2023.

\bibitem{zhu2023learning}
Y.~Zhu, Z.~Jiang, P.~Stone, and Y.~Zhu, ``Learning generalizable manipulation policies with object-centric 3d representations,'' \emph{arXiv preprint arXiv:2310.14386}, 2023.

\bibitem{gu2023rt}
J.~Gu, S.~Kirmani, P.~Wohlhart, Y.~Lu, M.~G. Arenas, K.~Rao, W.~Yu, C.~Fu, K.~Gopalakrishnan, Z.~Xu \emph{et~al.}, ``Rt-trajectory: Robotic task generalization via hindsight trajectory sketches,'' \emph{arXiv preprint arXiv:2311.01977}, 2023.

\bibitem{brohan2022rt}
A.~Brohan, N.~Brown, J.~Carbajal, Y.~Chebotar, J.~Dabis, C.~Finn, K.~Gopalakrishnan, K.~Hausman, A.~Herzog, J.~Hsu \emph{et~al.}, ``Rt-1: Robotics transformer for real-world control at scale,'' \emph{arXiv preprint arXiv:2212.06817}, 2022.

\bibitem{ingelhag2024robotic}
N.~Ingelhag, J.~Munkeby, J.~van Haastregt, A.~Varava, M.~C. Welle, and D.~Kragic, ``A robotic skill learning system built upon diffusion policies and foundation models,'' \emph{arXiv preprint arXiv:2403.16730}, 2024.

\bibitem{lee2024affordance}
O.~Y. Lee, A.~Xie, K.~Fang, K.~Pertsch, and C.~Finn, ``Affordance-guided reinforcement learning via visual prompting,'' \emph{arXiv preprint arXiv:2407.10341}, 2024.

\bibitem{rohmer2013v}
E.~Rohmer, S.~P. Singh, and M.~Freese, ``V-rep: A versatile and scalable robot simulation framework,'' in \emph{2013 IEEE/RSJ international conference on intelligent robots and systems}.\hskip 1em plus 0.5em minus 0.4em\relax IEEE, 2013, pp. 1321--1326.

\bibitem{kirillov2023segment}
A.~Kirillov, E.~Mintun, N.~Ravi, H.~Mao, C.~Rolland, L.~Gustafson, T.~Xiao, S.~Whitehead, A.~C. Berg, W.-Y. Lo \emph{et~al.}, ``Segment anything,'' in \emph{Proceedings of the IEEE/CVF International Conference on Computer Vision}, 2023, pp. 4015--4026.

\end{thebibliography}

\end{document}